\documentclass[letterpaper, 10 pt, conference]{ieeeconf}  

\usepackage[utf8]{inputenc} 
\usepackage[T1]{fontenc}    
\usepackage{hyperref}       
\usepackage{url}            
\usepackage{booktabs}       
\usepackage{amsfonts}       
\usepackage{nicefrac}       
\usepackage{microtype}      
\usepackage{xcolor}         

\usepackage{amsmath}
\usepackage[capitalise, nameinlink]{cleveref}

\usepackage{wrapfig}
\usepackage{graphicx}
\usepackage[normalem]{ulem}

\definecolor{mylightblue}{HTML}{89C5EC}  
\definecolor{mydarkblue}{HTML}{5753A3}  
\definecolor{mydarkorange}{HTML}{E37B3A}
\definecolor{mylightorange}{HTML}{FAAF6F}
\definecolor{myearth}{HTML}{7e6204}
\definecolor{snoopie}{HTML}{5FB177}

\definecolor[named]{JennGreen}{HTML}{21D19F}
\definecolor[named]{MaxBlue}{HTML}{79f4fc}

\definecolor[named]{TODORed}{HTML}{A30B37}

\newcommand{\algname}{\textsc{Snoopie}}

\definecolor{mygreen}{rgb}{0.12, 0.54, 0.30}
\definecolor{myred}{rgb}{0.78, 0.33, 0.37}
\hypersetup{
    colorlinks=true,
    linkcolor=snoopie,
    filecolor=magenta,      
    urlcolor=snoopie,
    citecolor=snoopie,
}

\IEEEoverridecommandlockouts                              

\title{\LARGE \bf
Will People Enjoy a Robot Trainer? \\ A Case Study with Snoopie the Pacerbot
}

\author{Maximilian Du$^{* 1}$, Jennifer Grannen$^{* 1}$, Shuran Song$^{1}$, Dorsa Sadigh$^{1}$%
\thanks{* denotes equal contribution. $^{1}$Stanford University. Correspondence to \texttt{maxjdu@stanford.edu}.}}

\begin{document}

\maketitle
\thispagestyle{empty}
\pagestyle{empty}


\begin{abstract}
The physicality of exercise makes the role of athletic trainers unique. Their physical presence allows them to guide a student through a motion, demonstrate an exercise, and give intuitive feedback. Robot quadrupeds are also embodied agents with robust agility and athleticism. In our work, we investigate whether a robot quadruped can serve as an effective and enjoyable personal trainer device. We focus on a case study of interval training for runners: a repetitive, long-horizon task where precision and consistency are important. To meet this challenge, we propose \algname{}, an autonomous robot quadruped pacer capable of running interval training exercises tailored to challenge a user's personal abilities. We conduct a set of user experiments that compare the robot trainer to a wearable trainer device -- the Apple Watch -- to investigate the benefits of a physical embodiment in exercise-based interactions. We demonstrate 
60.6\% better adherence to a pace
schedule and were 45.9\% more consistent across their running
speeds with the quadruped trainer. Subjective results also showed that participants strongly preferred training with the robot over wearable devices across many qualitative axes, including its ease of use (+56.7\%), enjoyability of the interaction (+60.6\%), and helpfulness (+39.1\%). Additional videos and visualizations can be found on our website: \href{https://sites.google.com/view/snoopie}{https://sites.google.com/view/snoopie}

\end{abstract}

\section{Introduction}

For health-conscious people and professional athletes alike, personal athletic trainers play a unique role in a fitness journey. The physical presence of trainers provides opportunities to not only give detailed feedback, but also offer intuitive demonstrations of proper techniques and drills. Across strenuous and repetitive exercises, a personal trainer also offers companionship and encouragement. 

In cases where a personal trainer is not available, wearable technologies have become a popular alternative. Their computational and sensory abilities endow them with some features of a personal trainer, including customized exercises and real-time feedback. However, these technologies cannot act in the environment of the user, leading to training challenges. Critically, the \textit{lack of physical embodiment} makes it difficult to demonstrate exercises and skills. Instead, these technologies passively react to the activities of a user. The feedback from screen displays or audio/haptic signals can also be less intuitive than following the lead of a trainer. 
To address these challenges of athletic training without a human personal trainer, we propose using technologies that are physically embodied, such as robots, as trainers. 
In recent years, robots have become more athletic. Quadrupeds, in particular, have become increasingly capable of all-terrain locomotion, payload carrying, and high-speed trotting \cite{chai_assessment_2021}. 

We hypothesize that the unique combination of physical embodiment and visuomotor reasoning abilities in robots will make them effective and enjoyable trainers for exercise. Unlike wearable devices, the physical embodiment of a robot allows it to act directly in the world, creating a tangible, embodied modality of interaction between user and technology. This tangible interaction also makes it easier to offer \textit{personalized} physical guidance. A user's abilities can be intuitively \textit{demonstrated} in an embodied manner by the user. Then, the robot can take the lead by \textit{guiding} the user through exercises at a challenging yet feasible intensity. In this way, physically-embodied robots can \uline{actively and adaptively guide}, rather than \uline{passively react}. Users might also see a physical robot as a companion that makes exercise more enjoyable \cite{kong_co-designing_2024, hwang_towards_2024, friedman_hardware_2003}.

\begin{figure}[t]
  \centering
\includegraphics[width=0.95\linewidth]{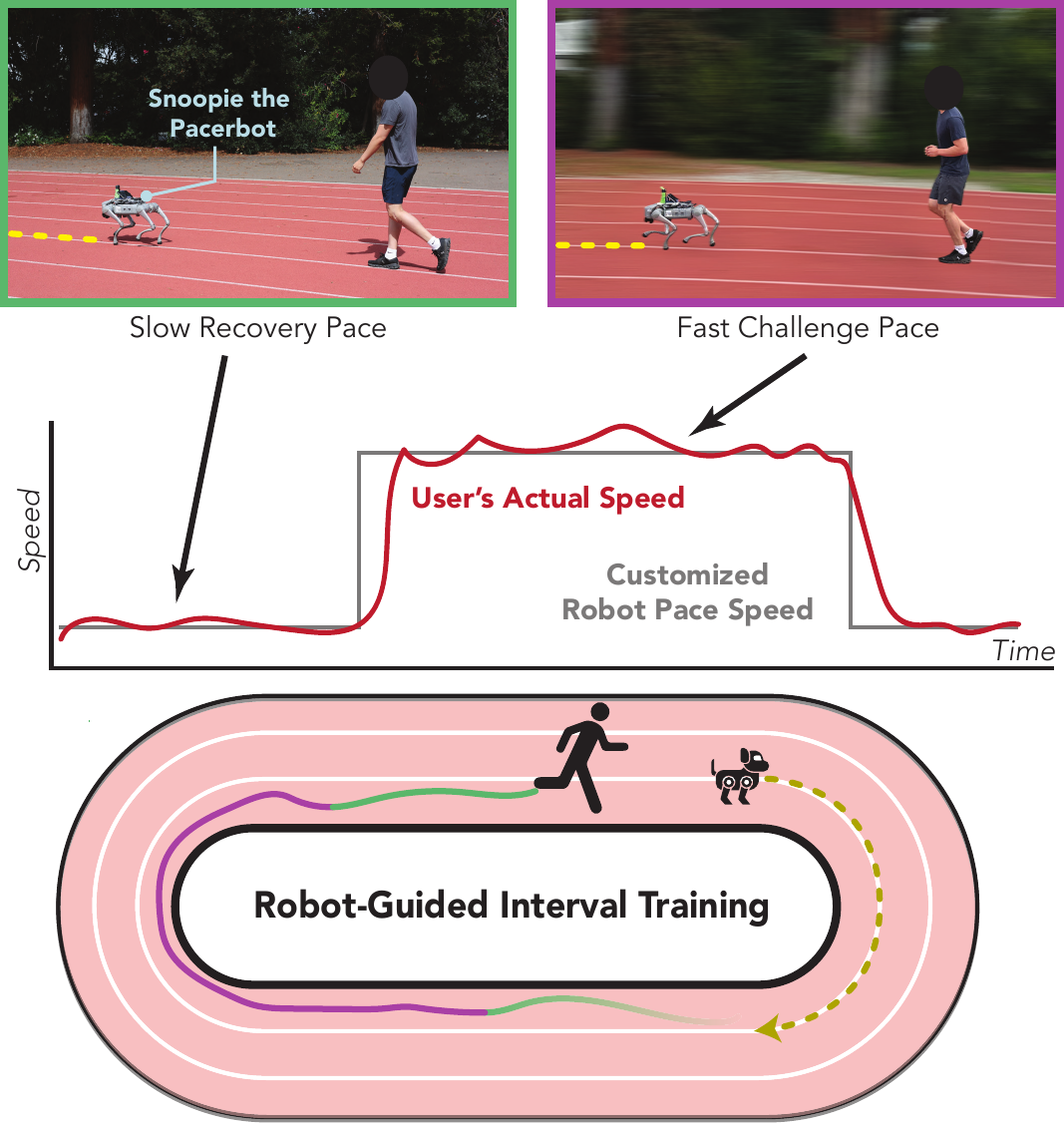}
  \vspace{-1mm}
  \caption{\textbf{Benefits of a Robot Trainer}. Much like a real personal trainer, \algname{} the Pacerbot leverages its embodied nature to guide runners at set paces that are customized to the runners' physical abilities. The quadruped's physicality leads to more accurate pacing exercises and greater perceived enjoyment compared to existing wearable technology.}
  \label{fig:pullfig}
    \vspace{-6mm}
\end{figure}



To test our hypothesis, we focus on a case study of applying robot quadrupeds for \textit{interval pace training} on a running track, which requires running at two different speeds for challenge and recovery (see \cref{fig:pullfig}). Training for effective pace control, while beneficial \cite{be_better_at_pacing}, 
is difficult for novice runners because they struggle to keep a consistent pace. 
Opportunistically, robot quadrupeds are reliable runners that are consistent in keeping pace (Section \ref{sec:robot_consistency}), making interval training a well-suited case study to see if users will find an embodied robot trainer effective and enjoyable. 

To tackle interval pace training, we present \algname{}, a modified Unitree Go2 robot quadruped capable of executing a custom fitness regimen involving precise speeds and time intervals. 
Using the user's demonstrated physical ability, the robot sets a target challenge pace and recovery pace to guide the user through a set of fast and slow interval exercises. 
To test the qualitative and quantitative impact of \algname{}, we conduct a user study comparing \algname{} to wearable pacing technology, and an extended interaction experiment investigating how \algname{}'s efficacy changes over multiple 20-minute sessions. 

We find that users with \algname{} achieved \textbf{60.6\%} better adherence to a pace schedule and were \textbf{45.9\%} more consistent across their running speeds. Critically, we found that users preferred training with the robot over wearable devices across many qualitative axes, including helpfulness (\textbf{+39.1\%}), fun (\textbf{+60.6\%}), and ease of use (\textbf{+56.7\%}). In an extended interaction setting, a participant exercising with \algname{} over 5 sessions was able to comfortably complete \textbf{11.0 kilometers} of jogging and walking, totaling \textbf{112.7 minutes}. 
Additional videos and visualizations can be found on our website: \href{https://sites.google.com/view/snoopie}{https://sites.google.com/view/snoopie}

\section{Related Works}

We review the related prior works on physically grounded human-robot interaction, with an emphasis on exercise tasks.


\subsection{Robot Exercise Trainers and Companions}
While much past work with physically-grounded human-robot interaction has focused on humans giving the robot feedback \cite{nikolaidis2014efficient, grannen2024vocal, grannen2025provox, matuszek2014gesture, lin2023giraf, arakawa2018dqn}, we consider the converse: robots demonstrating or teaching a human to do a task. 
Within the context of physical exercise training, humanoid-style robots have been used to encourage exercise via demonstration guidance -- prompting participants to mimic the robot's exercise poses \cite{avioz-sarig_robotic_2021, blanchard_technical_2022, fasola2013socially}. Other social robots also offer individualized corrective or encouragement feedback to enhance performance or healthy outcomes \cite{winkle2020couch, wang2024fitness, shen_artificial_2025, rossi2018socially, ros2016motivational}. Complementary to our approach, some of these works also offer abilities for the user to demonstrate their physical abilities or request customization \cite{tsai2022service, gross_roreas_2017, shen_artificial_2025}. In our work, we consider a more dynamic, outdoor application of track running and explicitly compare against alternative, non-embodied technology to test the importance of physical grounding. 

An alternative approach to physical robot assistance is exoskeletons, which supplement or give back physical abilities of their human users \cite{slade_personalizing_2022, zhang_human---loop_2017}. Like physical trainers, these exoskeletons can give customized assistance \cite{zhang_human---loop_2017, slade_personalizing_2022, tucker_preference-based_2020}. However, unlike physical trainers, these systems are meant to optimize the combined human-robot system. In contrast, the end goal of a robot physical trainer like \algname{} is to instill a physical ability solely in the user.  


\subsection{Wearable Technology for Training Runners}
Technology has a longstanding place in athletic training, and training runners is no exception. The most common technology assistance is wearable trackers that act as trainers by creating workouts, giving pace feedback, and even measuring statistics about the exercise like running form. Apple Watches \cite{apple_watch} and Fitbits \cite{fitbit} are examples of this general device class. Specialized apps like Strava can track a user's athletic progress and give feedback, including pace times, heart rate, and overall performance \cite{strava}. However, as previously discussed, these devices lack physical embodiment. Prior work has demonstrated the effectiveness of incorporating a physical presence in human-robot interaction settings \cite{bainbridge2008effect}, presenting an opportunity for improvement in this setting of training runners. In our experiments, we consider a physically embodied quadruped robot pacer, and compare against wearable technology. 


\subsection{Quadrupeds in Human-Robot Interaction}
The four-legged embodiment of quadrupeds makes them a popular choice of robots in human-robot interactions. In an assistive context, the all-terrain locomotion of quadrupeds have opened opportunities for guide dog replacements with blind and low-vision (BLV) users \cite{kim_understanding_2025, hwang_towards_2024}. These studies have revealed the unique and personalized needs for robot guide dogs, including transparency, interpretability, and natural communication \cite{kim_understanding_2025}. Our work is closely related to these BLV guidance projects, as we similarly focus on leading a human participant through a series of locomotion movements. Prior work demonstrating the success of quadrupeds in BLV settings highlights the potential of quadrupeds for pace-setting and guidance during locomotion.





The dog-like embodiment of robot quadrupeds has also inspired ongoing research into robots as companions \cite{hwang_towards_2024}. The Sony AIBO \cite{fujita_aibo_2001} was developed as a dog-like companion robot that learns and responds to feedback, leading to the perception of companionship \cite{friedman_hardware_2003}. While the AIBO is interactive, it is incapable of moving at speeds comparable to a human and thus does not consider more physically active exercise tasks. Larger, more physically capable quadrupeds have also shown companionship potential with neurodiverse individuals \cite{kong_co-designing_2024}, but they assume constant teleoperation and do not consider autonomous capabilities.
The embodied nature of the robot quadruped encourages a feeling of connection between robot and human \cite{hwang_towards_2024, kong_co-designing_2024}, which suggests promise for exercise-based interactions as well. 
We propose to explore this relationship with our interval running case study, where the criteria for companionship shift from social engagement and mutual responsiveness in play, to physical agility and robust and consistent autonomous behavior.

\section{Interval Running Pacing Setting}


\label{sec:problemsetup}
We consider the task of embodied exercise trainers with a case study in interval running pacing setting. We formalize our assumptions about this task setting.






\begin{figure*}[t]
  \centering
\includegraphics[width=0.9\linewidth]{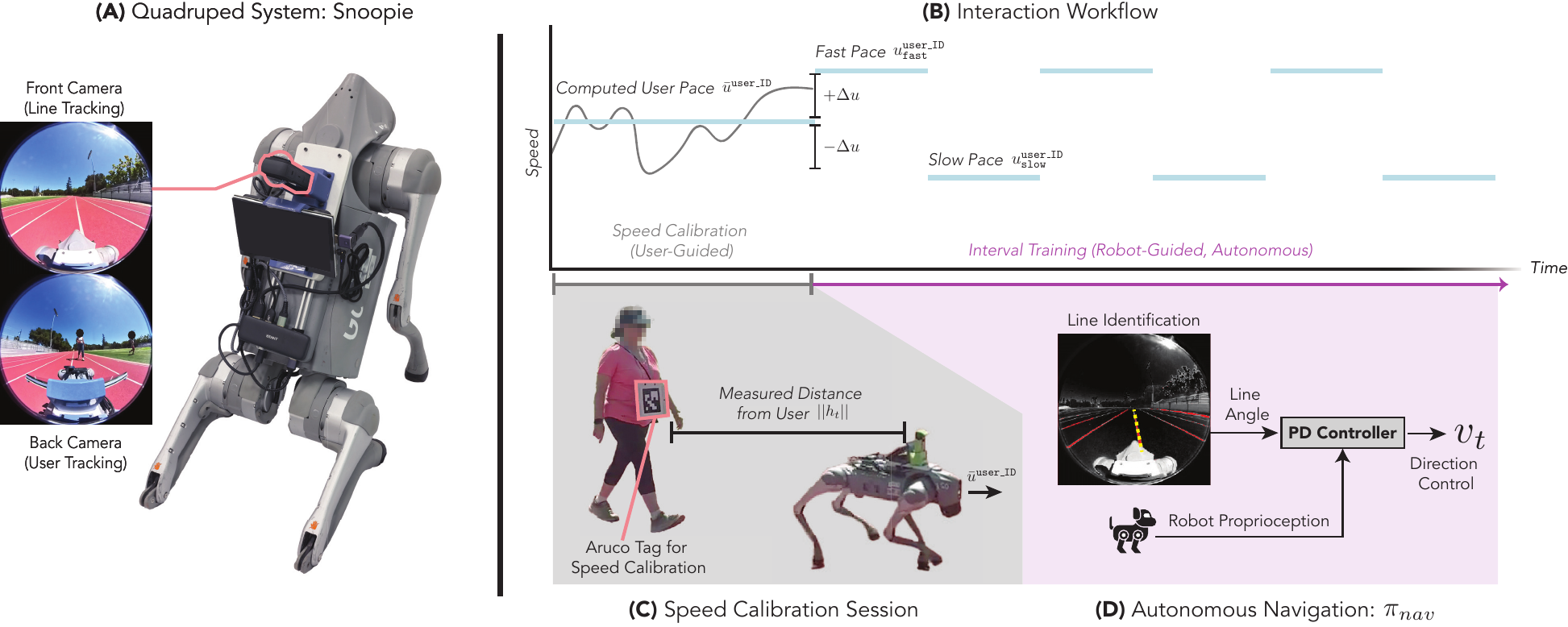}
  \vspace{-2mm}
  \caption{\textbf{Snoopie the Pacerbot}.  The pace-training robot is a Unitree Go2 quadruped fitted with two fisheye cameras for user interface and autonomous navigation \textbf{(A)}. The pace training interaction has two phases \textbf{(B)}: an initial speed calibration and the robot-led interval training exercise (\S \ref{sec:interaction}). During speed calibration \textbf{(C)}, the robot adjusts its speed to find a comfortable user pace. During interval training \textbf{(D)}, the robot alternates between a slow recovery pace and a fast challenge pace, all while navigating the race track autonomously.}
  \label{fig:methodsfig}
    \vspace{-5mm}
\end{figure*}

\subsection{Problem Setting and Assumptions}
We consider the problem of interval running pacing with a quadruped robot. We formulate each interval training session as a sequential decision making problem defined by components $( \mathcal{O}, \mathcal{A} )$. Each observation consists of RGB image frames $o_t \in \mathbb{R}^{H \times W \times 3}$ and the proprioceptive state of the robot $p_t \in \mathbb{R}^{3}$ that includes an orientation $\theta_t$, and 2D position $(x_t, y_t)$.
$\mathcal{A}$ is the action space for the robot containing velocity actions $v_t = \Delta p \in \mathbb{R}^3$. We assume a known mapping $\pi_l: v_t \rightarrow j_t$ from a desired velocity action vector $v_t$ to a set of robot joint positions $j_t$ to execute. In practice, this mapping is learned via reinforcement learning in simulation.  

To perform this task, we use a quadruped robot operating in a standard running track to Olympic specifications, including white-painted track lane lines against a brick-colored background. We define a standard $(x,y,z)$ coordinate frame in this setting relative to the robot position. We use two RGB cameras mounted on the robot with known intrinsics and extrinsics, which allows us to obtain a mapping $(o_{t,x}, o_{t,y})$ in pixel space to a coordinate $(x, y, z)$ relative to the robot.

We assume a training program schedule for interval running (similar to a Japanese 3x3 interval workout \cite{mckay_japanese_2024}) that segments each exercise session of length time $T$ into alternating slow and fast target speed segments ${T_{\texttt{slow}, 1}, T_{\texttt{fast}, 1}, T_{\texttt{slow}, 2}, T_{\texttt{fast}, 2}, ...} $. Each speed segment has a corresponding target speed $u_{\texttt{slow}}, u_{\texttt{fast}}$ determined by the exercise structure and calibration interactions with the user (\S \ref{sec:interaction})
The robot should lead this exercise by running in front of the user and traveling at these target speeds. 


\subsection{Autonomous Control for Pacing}
An interval training robot must bridge the capabilities between the low-level robot controller $\pi_l$ and the high-level running curriculum. We consider decomposing this problem into perception and control. Using observation images $o_t$ and calibration information from the onboard cameras, we integrate a perception model that identifies $N$ coordinate points $l_t \in \mathbb{R}^{N \times 3}$ of the track line in the robot coordinate frame. Then, we define a policy $\pi_{nav}: (u, l_t) \rightarrow v_t$ that takes these line positions and the exercise target speed $u \in \{u_{\texttt{slow}}, u_{\texttt{fast}}\}$ and converts them into target velocities $v_t$ for $\pi_l$ to ensure that the robot stays centered over the track line. In practice, $\pi_{nav}$ is implemented as PD controller with smoothing filters to encourage robustness to occlusion.

\subsection{Interaction Workflow}
\label{sec:interaction}

Robot physical trainers must also be able to adapt -- to assess the students' abilities and tailor their guidance to reasonably challenge these abilities through exercises \cite{poggensee_how_2021}. During the interval training, the robot runs at a slow recovery pace $u^{\texttt{user\_ID}}_{\texttt{slow}}$ and a fast challenge pace $u^{\texttt{user\_ID}}_{\texttt{fast}}$. These paces must be customized for each user's physical fitness and session-to-session motivation.


To determine $\{u^{\texttt{user\_ID}}_{\texttt{fast}}, u^{\texttt{user\_ID}}_{\texttt{slow}}\}$ for each user, the robot must understand the nominal pace of the user. To gain this understanding, we consider directly interacting with the user in a \textit{calibration session}. During this session, the robot starts running with a fixed initial pace $\hat{u}$, and observes the user's behavior from its camera as a coordinate point relative to the robot $h_t \in \mathbb{R}^3$. If the distance between the user and robot is widening (that is, if $||h_{t+1}|| > ||h_t||$), then the robot is running faster than the user's capabilities. Alternatively, if the gap is narrowing and $||h_{t-1}|| < ||h_t||$, the user is signaling that this pace is too slow. Using this continuous feedback, the robot trainer can converge upon a speed $\bar{u}^{\texttt{user\_ID}}$ that is comfortable for the user. The experience of this calibration session can also yield $\Delta u^{\texttt{user\_ID}}$, or the range of speeds that the user is willing to go. 

From $\bar{u}^{\texttt{user\_ID}}$ and $\Delta u$, we can calculate $u^{\texttt{user\_ID}}_{\texttt{fast}} \leftarrow \bar{u}^{\texttt{user\_ID}} + \Delta u$ and $u^{\texttt{user\_ID}}_{\texttt{slow}} \leftarrow \bar{u}^{\texttt{user\_ID}} - \Delta u$, which is executed in the predetermined time intervals through $\pi_{nav}$ and $\pi_l$. This combined system---the customized exercise, the navigational policy, and the low-level control policy---form the basis of an interval training track robot. 



\section{Snoopie the Pacerbot}
\label{sec:snoopie}
To complete the case study for interval training, we propose one implementation of the problem requirements (\S \ref{sec:problemsetup}). We present \algname{}: a robotic quadruped runner pacer.

\subsection{Hardware Setup}
\label{sec:hardware}
To meet the endurance and speed requirements of an interval training robot, we look at the Unitree Go2 quadruped. The Go2 comes with a trained $\pi_l$ that can run upwards of 2.6 m/s ($\approx$ 9.4 km/h) consistently, which is a challenging running pace for most novice runners \cite{running_times}. At this speed, the battery life surpasses one hour of continuous operation. The robot also runs at an accurate and consistent speed (\S \ref{sec:robot_consistency}). 
This combination of sufficient running speed, consistency, and endurance made the Unitree Go2 the ideal choice for our pacing experiments. 

To extract navigational and runner information from the environment, we get $o_t$ from a single Insta360 X4 camera. This device has a forward and backward-facing 180-degree fisheye camera, allowing perception of the entire area around the robot (\cref{fig:methodsfig}A). 




\subsection{Autonomous Navigation}
On a running track, the navigation policy $\pi_{nav}$ needs to follow the same track line consistently. The front-facing camera provides a wide-angle image that shows all of the track lines in the same frame. We use adaptive computer vision thresholding and contour detection from OpenCV \cite{opencv_library} to isolate the individual lines of the track, and find the principal axis of each detected contour to yield a set of line candidates. We select the line by leveraging the intuitive prior that the robot should follow the line which is horizontally-centered and closest to vertical.  A PID direction control signal is generated using the best line. Combined with the current speed interval $u$, this implementation of $\pi_{nav}$ outputs the velocities $v_t$ to the pretrained joint controller $\pi_l$.



\subsection{Pace Training Workflow}
\label{sec:system_adaptation}
As outlined in Section \ref{sec:interaction}, an interval pacing training robot must have a \textit{calibration session} before the exercise, which determines the interval speeds $u^{\texttt{user\_ID}}_{\texttt{slow}}$, $u^{\texttt{user\_ID}}_{\texttt{fast}}$. In our calibration session implementation we give users a wearable Aruco tag \cite{garrido-jurado_automatic_2014} that can be detected from the robot's back-facing camera (\cref{fig:methodsfig}A) as a measurement of distance $h_t$. While the robot navigates the race track, it increases speed in response to a decreasing $||h_t||$, and decreases speed in response to an increasing $||h_t||$.  At the end of calibration, we compute $\bar{u}^{\texttt{user\_ID}}$ by using the average calibration session speed, and we compute $u^{\texttt{user\_ID}}_{\texttt{slow}}$ and $u^{\texttt{user\_ID}}_{\texttt{fast}}$ by adding and subtracting $\Delta u = 0.5$ m/s, respectively. During experimentation, we found that setting a constant $\Delta u$ was a simpler implementation that heuristically balances an attainable challenge pace with a recovery pace that felt easy but not too slow. 



After the $u^{\texttt{user\_ID}}_{\texttt{slow}}$ and $u^{\texttt{user\_ID}}_{\texttt{fast}}$ paces have been calculated, the robot executes the interval running exercise (\cref{fig:methodsfig} D). In this phase, the robot runs at the set $u^{\texttt{user\_ID}}_{\texttt{slow}}$ and $u^{\texttt{user\_ID}}_{\texttt{fast}}$ speeds for preset intervals of time, set constant at $30$ seconds. In this phase, the robot is no longer adaptive to the user's speed and instead leads the user at the desired speed. During this phase, the users no longer wear the Aruco tag, allowing for more natural motion. For a visual representation of the two interaction phases, refer to \cref{fig:methodsfig}B.

\section{Experimental Evaluation}
\label{sec:experiments}

We evaluate the usability and perceived enjoyment of \algname{} as a running pacer in two settings: (1) a user study with $N=10$ novice runners comparing against existing, non-embodied trainer alternatives, and (2) a week-long deployment with a single participant. Both tasks consider a HIIT-inspired pacing exercise \cite{nike_hiit_interval} where runners alternate between intervals of a fast, challenge pace and a slow, recovery pace. All user studies are approved under Stanford University's IRB, with all participants providing informed consent.
All experiments were conducted on a standard 400-meter public running track with white lane markings conforming to Olympic specifications.

\begin{figure*}[t]
  \centering
\includegraphics[width=\linewidth]{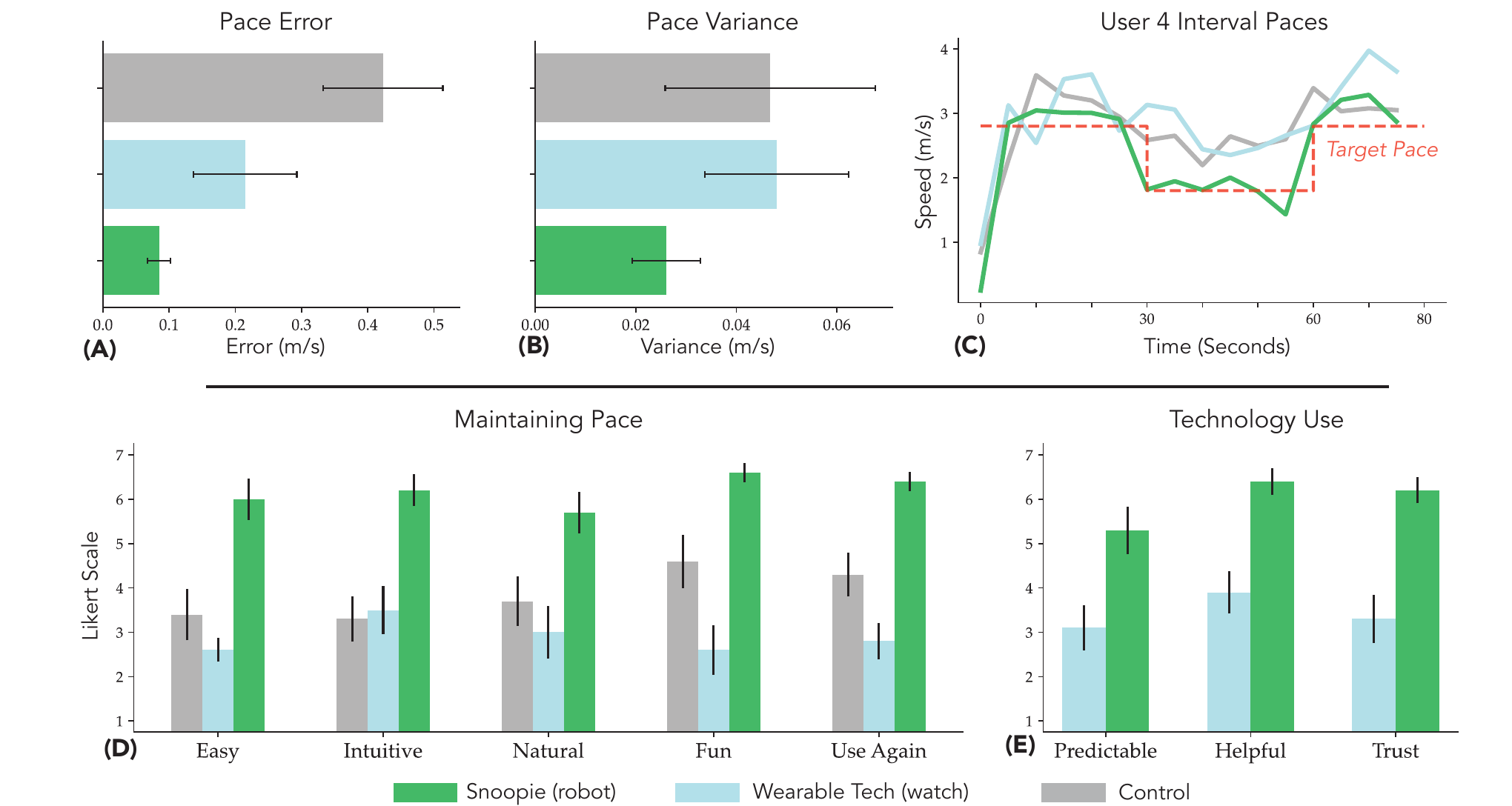}
  \vspace{-5mm}
  \caption{\textbf{User Study Results}. We track the participants' speeds during the pacing exercises to compute pacing performance and report the average and standard error across all three methods. On average, pacing with \algname{} led to lower error on target pace \textbf{(A)} and lower overall speed variance \textbf{(B)}. An example of a participant's data is shown on the right \textbf{(C)}. Notably, the user speed transitions during interval changes (indicated in red) are sharper when following the robot than with the \textsc{Wearable} and \textsc{Control} baselines. We also report the average and standard error of user ratings across eight subjective measures, finding that users also expressed a preference for \algname{} over wearable technology across all tested qualitative axes \textbf{(D, E)}.}
  \label{fig:user_study_quant}
    \vspace{-5mm}
\end{figure*}

\subsection{Verifying Robot Speed Accuracy and Consistency}
\label{sec:robot_consistency}
By using the robot as the interval pacer, we are assuming that it can travel at consistent and controllable speeds around the running track. To verify this, we measured the robot as it ran three representative paces of a novice runner demographic: $1.5$, $2.0$, and $2.5$ m/s. Across two trials of 30 seconds each per interval, we found that the Go2 achieves an average error of $0.026$ m/s from its target pace, with an average variance of $0.028$ m/s when aiming to keep a consistent pace. This accuracy and consistency demonstrated that \algname{} was suitable for the pacing exercise. 

\subsection{Technology Trainer User Study}
\label{sec:main_study}

In this study, each participant conducts a 30-second speed calibration phase and then completes four 30-second running intervals at alternating challenge (fast) and recovery (slow) paces. As described in Section \ref{sec:system_adaptation}, the challenge pace is 0.5\,m/s faster than the calibration pace, and the recovery pace is 0.5\,m/s slower than the calibration pace. We compared \algname{} with a commercially-available wearable exercise trainer -- the Apple Watch -- to investigate the differences between embodied and non-embodied trainer modalities.  

\smallskip
\noindent \textbf{Participants and Procedure}. We conduct a within-subjects study with 10 participants (4 female, 6 male, aged between 20 and 70). The participants were varied in their experiences with robots, with an average self-reported rating of $3.9 \pm 1.8$ out of $7$ . Most participants were not regular runners, with a self-reported score of $2.7 \pm 0.9$ out of $7$ for running frequency. Before the study, participants received an instruction sheet outlining the interval-based running task and the pacing capabilities of each technology: instantaneous speed feedback on the wearable pacer and the running speed of the robot. 


\smallskip
\noindent \textbf{Safety Considerations}. Safety and comfort of participants were a primary objective. The robot displayed its intended path and speed on a screen for transparency at all times. To alert the participant to speed changes, the robot gave audio indications leading up to the change. A proctor maintained line of sight to the robot and had override control access for the robot through a remote controller. Participants were also asked after the Calibration phase if they were comfortable with the speeds used in the pacing exercise.

\smallskip
\noindent \textbf{Independent Variables.}
We evaluate three pacing strategies for running: 1) \textsc{Snoopie} (robot quadruped), 2) \textsc{Wearable} technology pacer (Apple Watch), and 3) \textsc{Control} (no technology). In \textsc{Snoopie}, the participant follows the robot quadruped as it runs at the predefined target paces $(u^{\texttt{user\_ID}}_{\texttt{fast}}, u^{\texttt{user\_ID}}_{\texttt{fast}})$ through the three intervals. In \textsc{Wearable}, the participant is first informed of their target speeds. During the exercise, they have access to an Apple Watch displaying their instantaneous speed.  In \textsc{Control}, the participant is also informed of their target speeds, but they are unable to use technology to match these speeds. In both \textsc{Wearable} and \textsc{Control}, a study proctor indicates interval changes every thirty seconds. Each participant runs all three strategies, with randomized method ordering and naming. We do not test a human pacer baseline as it requires accurate self-pacing for the pacer, a confounding factor. 

\smallskip
\noindent \textbf{Dependent Measures}. We consider both subjective and objective measures. Throughout the study, participants wear a tracking device to log GPS coordinates during running. We use these coordinates to compute the average speed of the participant and compare it to the target speeds to yield an \textit{accuracy} measurement. We also compute the variance of the speed as a \textit{consistency} measurement. 

After completing each pacing condition, participants were asked to rate the method on several subjective dimensions about ease of operation (\textit{ease of use}, \textit{intuitiveness}) and enjoyability (\textit{naturalness}, \textit{fun}, and \textit{willingness to use again}). These dimensions of evaluation were selected based on the System Usability Scale~\cite{system_usability}. When applicable, participants also evaluated the technology pacer in terms of its \textit{helpfulness}, \textit{trustworthiness}, and \textit{predictability}. Participants were also encouraged to provide written feedback after the study.

\smallskip
\noindent \textbf{Hypothesis.} We hypothesized that:

\textbf{H1.} Participants being paced by \algname{} will have lower error (higher accuracy) when tracking the fast and slow interval speeds.

\textbf{H2.} Participants being paced by \algname{} will have lower within-interval variance (higher consistency). 

\textbf{H3.} Due to \textsc{Snoopie}'s physical embodiment, participants will find it more \textit{easy, intuitive, natural, fun, predictable, helpful}, \textit{trustworthy}, and likely to \textit{use again}. 

\smallskip
\noindent \textbf{Results -- Objective Metrics.} We present quantitative results for participant running accuracy (measured as pace error) \textbf{(H1)} and consistency (measured as speed variance) \textbf{(H2)} in \cref{fig:user_study_quant} (Top). Across all methods, \algname{} achieves the lowest pace error, indicating the highest accuracy \textbf{(H1)}. We attribute this to the robot’s physical embodiment and consistent pacing, which provides real-time visual cues of the target pace directly on the track. In contrast, the \textsc{Wearable} method requires participants to continuously monitor the watch and adjust their speed to match the displayed target pace. As expected, the \textsc{Control} condition exhibits the highest error, due to the absence of any speed feedback. 

Similarly, \algname{} exhibits the lowest pace variance, indicating the highest consistency \textbf{(H2)}. Interestingly, the \textsc{Wearable} method shows slightly higher variance than both the \algname{} and \textsc{Control} conditions. This is likely because participants must continuously adjust their speed to match the target displayed on the watch, which can introduce over- or under-corrections, whereas the robot provides direct visual cues on the track that are easier to follow. In contrast, the \textsc{Control} condition, while lacking feedback, shows relatively high variance that reflects the natural difficulty novice humans have in maintaining a consistent running speed, but without the additional fluctuations caused by constant speed corrections required in the \textsc{Wearable} condition.

We assess the significance of these results through one-way repeated-measures ANOVA tests. We find that interval training with \algname{} has a statistically significant effect on pace accuracy ($p=0.0127 < 0.05$). However, we find that the pace consistency results are not statistically significant, likely due to the natural variance in users' running paces.  

\begin{figure*}[t!]
  \centering
\includegraphics[width=0.9\linewidth]{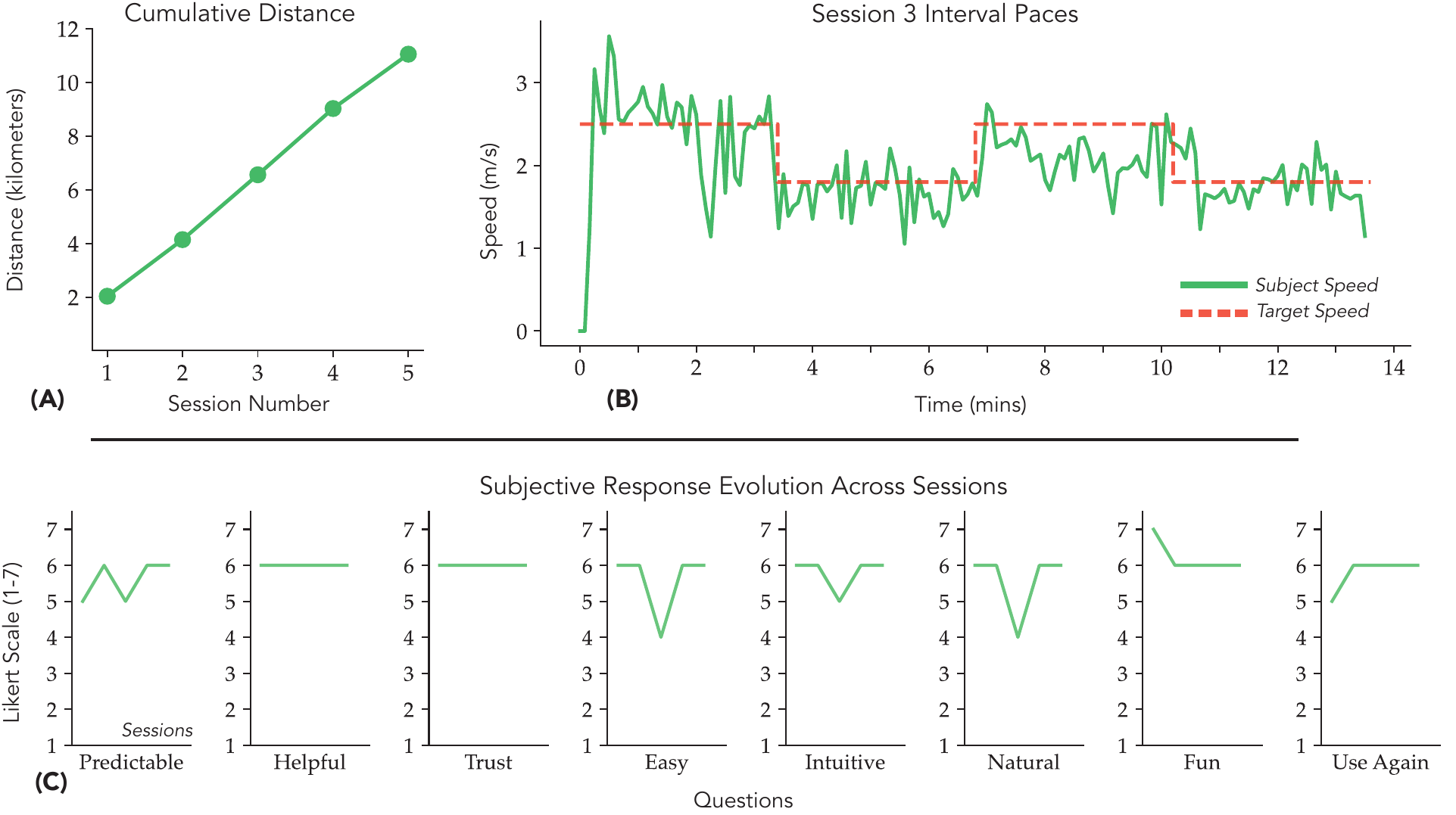}
  \vspace{-3mm}
  \caption{\textbf{Extended Interaction Session Results}. Across five sessions, the participant spent $112.7$ minutes with \algname{} and ran a total of $11.0$ kilometers (\textbf{A}). An excerpt of one session is displayed in (\textbf{B}), which shows his changing speeds as the robot brings him through the interval exercise. Throughout his extended interaction, his session-to-session qualitative ratings stayed relatively constant (\textbf{C}), a promising initial sign that \algname{}'s positive perception is not only due to novelty effects. }
  \label{fig:jensen}
    \vspace{-5mm}
\end{figure*}

\smallskip
\noindent \textbf{Results -- Subjective Metrics.} We present the qualitative survey results in \cref{fig:user_study_quant} (D, E), where participants rated each method on a 1–7 Likert scale \textbf{(H3)}. Overall, users preferred \algname{} across all question axes related to maintaining pace. Notably, they found the robot more \textit{fun}, \textit{intuitive}, and \textit{easy} to use (Figure \ref{fig:user_study_quant}D) \textbf{(H3)}. Participants also consistently indicated that they would \textit{prefer to use} \algname{} \textit{again} for paced interval training, citing the simplicity of keeping pace \textbf{(H3)}. One participant wrote, ``[it] felt very effortless to just focus on the dog and not managing [sic] my pace myself''.

Although the \textsc{Wearable} condition was quantitatively better than the \textsc{Control} condition at keeping pace, the non-physical embodiment of the Apple watch in the \textsc{Wearable} condition meant that the participants needed to look frequently at the watch screen while running, which was reported to be tedious, inaccurate, and uncomfortable. One stated that it ``felt like I was running either too fast or too slow'', and that they ``couldn't focus on running or enjoying it.'' As such, participants found the \textsc{Wearable} method less \textit{easy}, \textit{natural}, and \textit{fun} compared to \textsc{Control}. Across the three conditions, the \textsc{Wearable} condition was the least likely for participants to say they wanted to \textit{use again} (Figure \ref{fig:user_study_quant}D). 

When comparing the two technology pacers directly, users rated \algname{} as more \textit{predictable}, \textit{helpful}, and \textit{trustworthy} than the \textsc{Wearable} method \textbf{(H3)} (Figure \ref{fig:user_study_quant}E). In their written remarks, most users focused on the explicit embodiment of the robot. One user stated that using the robot ``doesn't require looking away from the track.'' Another reflected that ``I liked the dog the best because it felt like the least mental effort compared to the others.'' Finally, some users took the physical embodiment as a form of companionship, writing that ``the dog is cute'' and that ``it was fun to have someone (some dog) to accompany my running.'' In contrast, no user mentioned companionship with the wearable watch. 

Both objective and subjective metrics highlighted \algname{} as a better pacing approach. Importantly here, \uline{many of the perceived advantages of the \algname{} stem from its physical embodiment}. Users contrasted the intuitive following of the robot dog with the non-intuitive digital feedback of the wearable device. Some users even saw companionship in the robot as they ran with it, commenting about its visual appearance and the sound it made for speed change indication. 

We run one-way repeated measures ANOVA tests on these results and find that the results across all subjective measures are statistically significant, with the following $p < 0.05$ values: Ease ($p = 0.000044$), Intuitive ($p = 0.000347$), Natural ($p = 0.000542$), Fun ($p < 0.00001$), Use Again ($p <0.00001$), Predictable ($p = 0.005743$), Helpful ($p = 0.003481$), and Trust ($p = 0.002654$).

\subsection{Extended Interactions with \algname{}}
\label{sec:jensen_study}

While the user study in \cref{sec:main_study} demonstrated \algname's effectiveness in pacing participants enjoyably, we also sought to investigate whether this trend persists across multiple training sessions. Would \algname{} become burdensome or annoying over time? Are there benefits of \algname{} that become more apparent through extended use? 

Thus, we consider an extended interaction setting with a single participant (male, aged $26$, self-reported running frequency of $1$ out of $7$), who ran with the \algname{} pacer for 20–25 minutes per session across five sessions. Each speed interval lasted 3 minutes, alternating between a slow and fast pace for at least three sets per session. The interval lengths and alternating paces replicate a Japanese 3x3 interval workout \cite{mckay_japanese_2024}. The extended study user did not participate in the short-horizon user study. Each session began with the calibration phase to set their target paces for that session. We note this per-session calibration is important due to fluctuations with his comfort and energy levels from session to session. The participant completed the same post-study survey used in the main user study. Additionally, a wearable device tracked the participant's speed and position over all sessions.






\smallskip
\noindent \textbf{Results.} We report quantitative and qualitative results of the extended interaction experiment in \cref{fig:jensen}. Over five sessions, the participant ran a total of $11.0$ kilometers over a total of $112.7$ minutes of interval training with \algname{}, running at target speeds ranging from $1.3$ to $2.5$ m/s. We visualize a single representative session in \cref{fig:jensen} (B) where the participant aims to follow a target fast and slow speed of $2.5$ and $1.8$ m/s, respectively\footnote{Because $2.8$ m/s is beyond the robot range, we cap the speed at $2.5$ m/s.}. While natural variance appears in the participant’s pace during the longer training sessions, we observe that he consistently and rapidly adapts to changes by aligning with the robot’s demonstrated pace. 

Critically, we highlight the qualitative results from this study, shown in \cref{fig:jensen} (C). After each session, the user responded to the same survey questions, enabling us to examine potential trends as his exposure to \algname{} increased. Most measures -- such as ease, intuitiveness, naturalness, helpfulness, and trust -- remained relatively stable across sessions, indicating that \algname{} did not become an increasing burden over the extended interaction. We did observe a slight decline in perceived ``fun'' after the first session, suggesting a possible novelty effect for the initial interaction. Conversely, ratings of “would use again” showed a slight increase over time, which may reflect initial skepticism or unfamiliarity that diminished with repeated use. 

In the survey, the participant also commented about his relationship with \algname{}, as well as additional companionship features that emerge from its physical embodiment. 
%
The participant remarked that ``running with the [robot] seems to make time pass a bit faster. It doesn't feel like I run as far as I actually do, which is good.'' Reflecting on the robot's embodied guidance, the participant wrote, ``[the] physical entity guide... provided good motivation to push myself a little and have some objective to aim for.'' 
These remarks focused on the robot dog as a physical entity capable of encouragement or distraction from a strenuous task
Although it would take more extended interactions across a more diverse audience to make stronger conclusions about companionship, this single extended interaction experiment provides a promising sign that \algname{}'s physical embodiment serves a purpose beyond accurate pace setting in the context of physical training.




\section{Discussion}
In this work, we consider the question: does the physical embodiment of robots make them more effective and enjoyable trainers? To study this question, we propose a case study of interval pacing exercises, which we accomplish with \algname{}, a novel application of robot quadrupeds as personal trainers for this task. In our $N=10$ user study, we find that users are more accurate (\textbf{+60.6\%}) and consistent (\textbf{+45.9\%}) with \algname{} compared to wearable technology. Critically, users rate \algname{} higher across all tested qualitative attributes, especially ease of use (\textbf{+56.7\%}), fun (\textbf{+60.6\%}), and helpfulness (\textbf{+39.1\%}).

\smallskip

\noindent \textbf{Takeaway -- Calibration.}
This case study also surfaced a unique insight about robot trainers -- in order to lead effectively, the robots must also be calibrated to a users capabilities. Thus, not only must the robots influence the person by guiding or demonstrating, the person must first establish a calibration with the robot to form a shared grounding about the goals for the session. We instantiate this idea with a two-step interaction structure for each session: an initial calibration routine (human leader) followed by a training routine (robot leader). This insight can be applied beyond interval pacing to other athletic activities as well, for example, demonstrations in yoga. 

\smallskip

\noindent \textbf{Takeaway -- Companionship.} 
In addition to improving pace accuracy and usability, we found that participants often described the robot as a motivating companion during training. Unlike wearable devices that deliver abstract metrics, the quadruped’s embodied presence created a sense of ``running with'' someone rather than training alone. This feeling of companionship was shaped by subtle anthropomorphism -- users ascribed social and motivational qualities to the robot despite its limited interaction capabilities. In contexts such as play, anthropomorphism often leads people to treat robots as partners or pet-like figures; in exercise, however, this tendency takes on a more functional form, where the robot is perceived as a training partner who both demonstrates and motivates. This insight can inform future design decisions for embodied trainers, suggesting that emphasizing demonstration and shared activity -- rather than relying solely on feedback or critique -- may better sustain engagement.

\smallskip

\noindent \textbf{Limitations and Future Work.} While \algname{} focuses on interval training for running, the broader concept of leveraging a robot's embodiment for exercise training is widely applicable, as demonstrated in prior work ~\cite{avioz-sarig_robotic_2021, winkle2020couch}. \algname{} is currently limited to race-track settings as \algname{}'s path planning capabilities rely on following track lines. 
Extending the system to operate in more general running settings such as sidewalks would significantly expand the scope of possible interactions and enable easier use over long-term training.
In these longer-term interactions, we plan to consider different levels of user-trainer adaptation beyond the calibration phase, for example following a training schedule and adjusting goals based on user feedback.
\section*{Acknowledgments}
\footnotesize

\noindent This work was supported in part by NSF Award 2143601, 2037101, 2132519, 2132847 and 1941722, ONR YIP, and ONR MURI N00014-25-1-2479. 

\noindent \textit{AI Tool Disclosure}: The authors are responsible for all content in this article. AI tools (Claude, ChatGPT) were used in a limited capacity (minor grammar enhancement and code autocompletion). 


\bibliographystyle{IEEEtran}
\bibliography{references}

\end{document}